# Classification of Multiple Diseases on Body CT Scans using Weakly Supervised Deep Learning

**Original Research**

Fakrul Islam Tushar[a,b], Vincent M. D'Anniballe[a], Rui Hou[a,b], Maciej A. Mazurowski[a,b], Wanyi Fu[a,b], Ehsan Samei[a,b], Geoffrey D. Rubin[c], Joseph Y. Lo[a,b].

[a]Center for Virtual Imaging Trials, Carl E. Ravin Advanced Imaging Laboratories,
  Department of Radiology, Duke University School of Medicine, Durham, NC 27705 USA.
[b]Department of Electrical and Computer Engineering, Duke University, Durham, NC 27705 USA.
[c]Department of Medical Imaging, University of Arizona, Tucson AZ.

**Summary**:

A rule-based algorithm enabled automatic extraction of disease labels from over 10,000 radiology reports; these weak labels were used to create deep learning models to classify multiple diseases for three different organ systems in body CT.

**Key points:**

- Labels extracted by the rule-based algorithm were 91% to 99% accurate by manual validation.

- Deep learning models developed for each of three organs systems, consisting of *(a)* lungs and pleura, *(b)* liver and gallbladder, and *(c)* kidneys and ureters, enabled the classification of multiple diseases with areas under the receiver operating curve ranging from 0.65 to 0.97, 0.62 to 0.89, and 0.68 to 0.92, respectively.

# ABSTRACT

**Purpose**

To design multi-disease classifiers for body CT scans for three different organ systems using automatically extracted labels from radiology text reports.

**Materials & Methods**

This retrospective study included a total of 12,092 patients (mean age 57 ± 18; 6,172 women) for model development and testing (from 2012-2017). Rule-based algorithms were used to extract 19,225 disease labels from 13,667 body CT scans from 12,092 patients. Using a three-dimensional DenseVNet, three organ systems were segmented: lungs and pleura; liver and gallbladder; and kidneys and ureters. For each organ, a three-dimensional convolutional neural network classified no apparent disease versus four common diseases for a total of 15 different labels across all three models. Testing was performed on a subset of 2,158 CT volumes relative to 2,875 manually derived reference labels from 2133 patients (mean age 58 ± 18;1079 women). Performance was reported as receiver operating characteristic area under the curve (AUC) with 95% confidence intervals by the DeLong method.

**Results**

Manual validation of the extracted labels confirmed 91% to 99% accuracy across the 15 different labels. AUCs for lungs and pleura labels were: atelectasis 0.77 (95% CI: 0.74, 0.81), nodule 0.65 (95% CI: 0.61, 0.69), emphysema 0.89 (95% CI: 0.86, 0.92), effusion 0.97 (95% CI: 0.96, 0.98), and no apparent disease 0.89 (95% CI: 0.87, 0.91). AUCs for liver and gallbladder were: hepatobiliary calcification 0.62 (95% CI: 0.56, 0.67), lesion 0.73 (95% CI: 0.69, 0.77), dilation 0.87 (95% CI: 0.84, 0.90), fatty 0.89 (95% CI: 0.86, 0.92), and no apparent disease 0.82 (95% CI: 0.78, 0.85). AUCs for kidneys and ureters were: stone 0.83 (95% CI: 0.79, 0.87), atrophy 0.92 (95% CI: 0.89, 0.94), lesion 0.68 (95% CI: 0.64, 0.72), cyst 0.70 (95% CI: 0.66, 0.73), and no apparent disease 0.79 (95% CI: 0.75, 0.83).

**Conclusion**

Weakly-supervised deep learning models were able to classify diverse diseases in multiple organ systems.

# I. INTRODUCTION

Artificial intelligence has the potential to improve disease diagnosis, exemplified by the widespread development of machine learning algorithms for classification, segmentation, and detection tasks (1-4). Despite the potential, most disease classification algorithms target only a single disease making their scope substantially narrower than that of clinical practice (5, 6). Currently, the major bottleneck for developing multi-disease artificial intelligence algorithms is a lack of annotated data. Training these algorithms has traditionally relied on manually labeling large numbers of medical images, requiring rigorous standardization, clinical expertise, and a substantial time commitment (7). The manual labeling of a dataset of radiographs can take substantial amounts of time, and the labeling of large datasets of cross-sectional imaging studies can become impractical.

Ample training data already exist in electronic health records. Unlike supervised learning, which relies on well-annotated data, weakly-supervised learning allows the use of incomplete or inaccurate data. Although less precise than hand labeling, weak supervision enables the systematic processing of vast amounts of disease information from existing radiology text reports by automatically labeling medical images. Such automatic labeling may mitigate the need for any further human annotation. This approach has already yielded several large medical imaging datasets (8-10). For CT specifically, DeepLesion is a dataset developed by extracting over 32,000 lesions in multiple organs from the radiology reports associated with over 4,427 unique patients (11-14). In other studies using report text for weak labeling, Draelos et al (15) annotated up to 83 thoracic abnormalities from 19,993 unique patients. 36,000 chest CT volumes, and Eyuboglu et al (16) annotated up to 26 anatomical regions in over 6,000 whole body fluorodeoxyglucose-PET/CT exams. These studies demonstrate the feasibility of weakly supervised annotation of large datasets to enable deep learning in medical imaging.

Compared to prior works, this study demonstrates the feasibility of weak supervision to create multiple, organ-specific models that provide multi-disease classification of body CTs based on existing radiology reports. Body CTs of the chest, abdomen, and pelvis were selected because they are performed commonly and encompass a variety of organs and diseases within a large portion of the body. A rule-based algorithm was chosen to annotate reports for multiple disease labels because it can achieve high accuracy and is readily adjustable to address new or expanded tasks. Three organ systems were targeted: the lungs and pleura, liver and gallbladder, as well as kidneys and ureters. These systems were chosen because they represent large structures with very different anatomical appearance, location, and have a range of disease manifestations. Using the corresponding annotated reports, a separate convolutional neural network (CNN) was trained for each organ system to classify CTs as being positive for one or more diseases or no apparent disease. It was hypothesized that a weak supervision framework based on a radiology report text would allow a CNN model to classify multiple diseases, even without information about the disease location and despite the heterogeneous appearance of each organ and disease. This work demonstrates the feasibility of creating organ-specific multi-disease classifiers for CT.

## II. MATERIALS AND METHODS

**Study Design and Overview**

Institutional Review Board approval was obtained, and informed consent was waived for this retrospective study that was compliant with the Health Insurance Portability and Accountability Act. Our earlier work in CT disease classification was trained on much smaller subsets of fewer than 1,600 volumes to classify a single disease of the chest (17). A related study performed classification of multiple chest diseases but was limited to non-contrast chest CTs (15). Compared to our previous study on automated labeling (18), the rule-based algorithms were refined and the number of patients were increased in the present study.

**Disease Labeling and Dataset Mining**

Labels were extracted from the free-text radiology reports using a rule-based algorithm to label body CT studies which were performed between 2012 and 2017 (18). A total of 414,438 reports were initially included. Reports were excluded if they were incomplete or duplicates ($n$ = 80,130) or if they were non-body radiology reports ($n$ =73,079). In total 261,229 reports from 112,501 patients were included. Among these studies, 165,659 reports were used to extract labels from the lungs and pleura from 74,944 patients, 96,532 reports for liver and gallbladder from 50,086 patients, and 87,334 reports for kidneys and ureters from 46,527 patients (Figure 1).

In this dataset, the sum does not correspond to the total number of volumes or patients reports, because one report could have multiple diseases in one or more of the three organ systems. Moreover, reports were selected separately for each disease and organ combination, resulting in different numbers for each organ that represent the natural prevalence in our study population. Chest CT protocols were excluded from analysis of abdominal organs (liver and gallbladder, kidneys and ureters). Only the findings section of the report was used to limit labels to image-based information and to minimize the influence of other factors such as scan indication or previous medical history. The findings section was tokenized into sentences and processed in series by a set of rules. The algorithm is described in the code repository (https://gitlab.oit.duke.edu/railabs/LoGroup/multi-label-weakly-supervised-classification-of-body-ct).

The rules were built upon general relationships, which can be applied to different organs by adding keywords to the dictionary. For example, *if* a sentence contained the organ name "liver" *or* "hepatic" *and* abnormality "lesion," *and* without any negation keywords such as "no" or "without," *then* the label was positive for liver lesion. By substituting other organ names such as "kidney" or "renal," the same rule would apply to kidney lesions. By using approximately 30 rules and 500 keywords, this method could be applied to different organs and diseases, such as renal atrophy or pleural effusion. The labels were considered "weak" for two reasons: 1) they were applied to the entire volume without disease localization, and 2) the rule-based decisions included some expected imperfections due to its automated implementation on existing report data.

This dataset represents a typical clinical sample, in which patients may undergo multiple scans or have multiple positive findings. Furthermore, this sample encompassed CT exams from a large health system that was not limited to any acquisition or reconstruction type. The lungs and pleura were labeled as no apparent disease versus having one or more of four diseases: atelectasis, nodule, emphysema and/or effusion. The liver and gallbladder were labeled as no apparent disease or hepatobiliary calcification, lesion, dilation, and/or fatty. The kidneys and ureters were labeled as no apparent disease or stone, lesion, atrophy, and/or cyst. These abnormalities were selected based on their high prevalence and varied appearance. Each abnormality or "disease" may comprise multiple keywords that were grouped together based on similarity of language or overlap in image appearance. For example, lung nodule and mass were grouped into the nodule class; biliary calculus, calcification, and gallstone to hepatobiliary calcification class; and fatty liver and steatosis to fatty liver class. The rule-based algorithm labeled a report as "normal" only in the absence of dozens of other abnormal labels (e.g., spiculated, surgically absent) that were not otherwise analyzed in this study. A comprehensive list of all abnormal labels can be found in the code repository associated with this study. If the disease labeling pipeline was unable to definitively categorize a report as positive for disease or no apparent disease (e.g., there was no mention of the organ system), then the report was labeled as uncertain and was not included in this study.

Figure 1 also shows how the rule-based algorithm distilled the larger number of patients down to 19,255 volume-level labels in 13,667 CT volumes from 12,092 patients. For each organ, the image and report data were randomly divided by patient into subsets to train (70%), validate (15%), and test (15%) the model. To establish the reference standard for labeling, reports for patients in the test set were manually assigned disease labels by a graduate student with gross anatomy training and 1 year of experience (V.M.D.) supervised by a board-certified radiologist with 30 years of experience (G.D.R.). The rationale for this reference standard was that a smaller representative subset of manually labeled reports was practicable, whereas labeling all 165,659 reports would have been prohibitive in both cost and time.

**Image Acquisition**

The CT scanners were either General Electric (Revolution, 750 HD, or VCT) or Siemens (Flash or Force). Continuous section thickness was between 0.625 and 1.25 mm. The contrast agent was between 75 and 150 mL of Isovue 370 depending upon the clinical protocol and patient size.

**Image Segmentation**

A segmentation model was used to guide extraction of organ-specific patches from CT volumes for subsequent classification. Figure 2 shows the overall classification pipeline used in this study. The DenseVNet (19) segmentation model was trained with the four-dimensional extended cardiac-torso (XCAT) dataset (20), which contains CT volumes and corresponding multi-organ segmentation masks. The 50 XCAT volumes from 50 unique patients were randomly assigned into 44 training and 6 validation. Because the XCAT training set contains normal anatomy only, segmentation errors were observed when the normally aerated lungs were replaced by disease. The segmentation model was fine-tuned using 30 additional, randomly selected, diseased lung volumes (10 edema, 10 atelectasis, seven pneumonia, and three nodule). The resulting segmentation masks were then manually corrected to the lung margins. These 30 diseased volumes with corrected segmentation were combined with 10 normal XCAT training volumes and used to fine-tune and produce the final segmentation model. The initial segmentation model performed well on both normal and diseased liver and kidneys volumes, so no further fine-tuning of the segmentation model was performed for those two organs. Organ masks were not available for the gallbladder and ureters, so those were included within expanded patches that contained the liver and kidneys respectively as described in the next section.

**Pre-processing and Weakly Supervised Image Classification**

Prior to classification, all CT volumes were resampled to voxels of size 2 mm × 2 mm × 2 mm by B-spline interpolations, clipped to intensity range (-1000, 800) HU for lungs and pleura, (-200, 500) HU for liver and gallbladder, and kidneys and ureters, and normalized to 0 mean and 1 standard deviation. To reduce computational expense and the influence of background organs, three-dimensional (3D) patches were placed around each organ: 224×160×160 (Z×W×H) for lungs and pleura and 96×128×128 (ZxWxH) for both liver and gallbladder, as well as kidneys and ureters. Patch sizes were based on organ size plus a margin to allow for patient variability and to include most or all of the gallbladder in the liver and gallbladder, as well as ureters in the kidneys and ureters systems. The segmentation module was used to guide individualized placement of patches, and kidney patches were offset anteriorly to remain within the body.

The architecture of the 3D CNN used in this study was inspired by Resnet (21). As shown in Figure 2b, features were learned in three resolution scales. After each resolution, the features were halved by max-pooling, and the number of filters was doubled. After the third resolution, the last R-block features were passed through batch-normalization, rectified linear unit, global max-pooling, dropout, and finally sigmoid classification layer for the multi-label predictions. The Adam optimizer was used to optimize the weights, and weighted cross-entropy was used as the loss function. The uniform distribution did initialization of the weights. In order to retain the natural prevalence and co-occurrence of diseases, no class-balancing was performed during training.

Training took approximately 26 hours for segmentation (Python TensorFlow v1.5) and 40 hours for classification (Python TensorFlow v2.0) using 4 TITAN RTX GPUs (Nvidia Corporation, Santa Clara, CA). All models' weights, initial hyper-parameters, and code are publicly available (https://gitlab.oit.duke.edu/railabs/LoGroup/multi-label-weakly-supervised-classification-of-body-ct). The 13,667 CT volumes and labels will also become publicly available pending institutional approval.

*Statistical Analysis*

Performance was assessed using the receiver operating characteristic (ROC) area under the curve (AUC). The 95% CIs were calculated by using the DeLong method as implemented by the package pROC 1.16.2 in R 3.6.1 with 2000 bootstrapping samples (22).

### III. RESULTS

#### Patient Overview

Volume-level labels for this study were generated by the rule-based algorithm for 19 255 findings in 13 667 CT volumes from 12092 patients. Figure 3 shows the total dataset and distribution of labels, volumes, and distinct patients among the three targeted organs.

The study population was from a health system composed of multiple hospitals. The average age of the patients was 57 ±18 years; the median age was 61 years. The overall percentage of women was 51% (6,172 of 12,092). Among the 13,667 CT volumes, 72% (9,836 of 13,667) were with contrast agent and 28% (3,831 of 13,667) were without contrast agent. The 13,667 CT volumes consisted of five protocols: 5,099 (37%) chest-abdomen-pelvis, 5,085 (37%) abdomen-pelvis, 2,886 (21%) chest, 488 (4%) abdomen, and 109 (1%) chest-abdomen. There was no exclusion performed based on age, scanner equipment or protocols, contrast agent, or type of reconstruction. reconstruction.

Table 1 shows the number of patients and volumes used in classification tasks for each organ system. For each of the three organ systems, the rule-based algorithm, labeled each volume as no apparent disease versus having one or more of four diseases. Figure 4 illustrates the co-occurrence and association between diseases for this multi-label data set. The test subset consisted of 2,158 (lungs and pleura = 771; liver and gallbladder = 652; kidneys and ureters = 749) CT volumes from 2133 distinct patients (mean age 58 ± 18; 1079 women) with 1,154 labels for lungs and pleura, 787 labels for liver and gallbladder, and 934 labels for kidneys and ureters.

**Label Extraction and Classification Models**

When compared to manually validated labels, the 2875 rule-based algorithm derived labels in the test set were identified with accuracy from 91% to 99% and F-score from 0.85 to 0.98. Table 2 displays the labeling accuracy for each organ system and disease class. Figure 5 shows the performance of the multi-label classification models for the lungs and pleura, liver and gallbladder, as well as kidneys and ureters, with ground truth based on the manually obtained test set labels.

In general, the classification performance was higher for diffuse abnormalities compared to focal abnormalities. For the lungs and pleura, AUC was greater than 0.80 for the diffuse lung diseases of effusion and emphysema, moderate for atelectasis with AUC of 0.77 (95% CI: 0.74, 0.81), but poor for nodules with AUC of 0.65 (95% CI: 0.61, 0.69). For liver and gallbladder, fatty liver and dilation demonstrated the highest performance with AUC of greater than 0.80, liver lesion was moderate at 0.73 (95% CI: 0.69, 0.77), while the hepatobiliary calcification class performed poorly with less than 0.70. For the kidneys and ureters, performance was good at AUC of greater than 0.80 for kidney stone and atrophy classes, and moderate for kidney lesion (0.68, 95% CI: 0.64, 0.72) and cyst (0.70, 95% CI: 0.66, 0.73) classes. Examples of images that were classified shown in Figure 6.

## IV. Discussion

The main purpose of this work was to create a weakly supervised 3D classification workflow that can be generalized to many diagnostic tasks in body CT. To test this workflow, classifiers distinguished between multiple diseases or no apparent disease among three different organ systems: lungs and pleura, liver and gallbladder, as well as kidneys and ureters. The organ systems and diseases were intentionally chosen to represent a wide variety of locations and appearances. Unlike conventional supervised learning, the proposed system used automated rules to analyze radiology text reports and avoid radiologist annotation efforts. This approach provides a form of weak supervision since each volume-level label (e.g., lung nodule) applies to an entire organ, although the abnormality may be present in only a portion of the volume.

The performances reported in this study do not match that of traditional, fully supervised studies where individual findings are annotated. However, weak supervision offers several advantages. By using automated label extraction, a vast data set from a large health system was able to be efficiently labeled. Such a scalable approach allows better representation of diverse patient populations, scanner equipment and protocols, as well as organs and diseases. In contrast, manually curating and labeling such a large, diverse data set would be prohibitive.

Many of the existing studies in medical image artificial intelligence have focused on a single abnormality or organ system (1, 4-6, 9, 23-25). This study expands on prior works by gathering a diverse dataset with 15 possible labels (12 abnormalities, and three no apparent disease labels for three different organ systems) from a large data set of over 13,000 body CT scans of over 12,000 distinct patients. Unlike common binary tasks such as disease presence versus absence, performing multi-label classification for several co-occurring diseases is exponentially more challenging. Furthermore, classification of multiple diseases and organs may have a different clinical application. Whereas a single disease model (eg, lung nodule detection) targets radiologist performance for that disease, a broader approach that encompasses multiple diseases and organs may instead seek to improve workflow, such as to conduct computer-aided triage for volumes with the highest or lowest likelihood of suspicion. Similar studies, such as Wang et al (9), explored a multi-disease classifier for eight common thorax diseases in two-dimensional chest radiography, and Draelos et al (15) classified CTs with 83 abnormalities in the chest. While those studies were limited to the chest region, the present study extended this weak-supervision approach to other organs in the abdomen and pelvis. The lung multi-class classification results presented here are most comparable to the Draelos et al (15) CT-Net-9 multi-label model trained with nine abnormalities: nodule 0.65 versus 0.68, atelectasis 0.77 versus 0.68, and pleural effusion 0.97 versus 0.94 for our model versus CT-Net-9, respectively.

Compared to two-dimensional CNNs, working with 3D CNNs is computationally very expensive, especially for large CT volumes. For this reason, the data was downsampled to voxels of size 2 mm $\times$ 2 mm $\times$ 2 mm to allow a reasonable balance between batch versus patch sizes. This reduced resolution may account for the lower performance for focal disease like lung nodules, compared to diffuse disease like atelectasis and effusion. This trend was also observed for the liver and gallbladder, as well as kidneys and ureters. The model had higher performance on diffuse diseases (such as fatty liver or renal atrophy) compared to focal diseases (such as gallstones and kidney lesions). There were a few notable exceptions of focal disease that had high AUC values (> 0.80), such as for kidney stone classification. This result likely reflected the relative ease of detection because the CT scans for kidney volumes tended to be from specialized kidney stone protocols.

This study has several limitations. This was a retrospective study from a single institution, and although the health system is composed of multiple hospitals, the patient population largely draws from a single geographic region, and reporting tendencies are likely to overlap amongst radiologists from the same department. Although this study encompassed multiple organs, the selected three sets of organ systems were independently processed for disease classification. This simplified the challenges that could be imposed by the interactions of multiple organs. Moreover, only a few of the most common abnormalities were considered, but results could have been confounded by co-occurrence of other diseases that were not otherwise analyzed. Finally, there were inevitable errors in the rule-based labeling due to the complexity of sentences, diversity of expression in free-text narration, and typographical errors. Literature suggests that 2-20% of radiology text reports are estimated to contain demonstrable errors (26). However, studies have shown that such errors can be ameliorated with large amounts of data (27). Other studies have used natural language processing that may provide more accurate analysis of the report text (28-30).

Future work will investigate important data extraction issues, such as the relationship of CT resolution on classification performance, particularly stratified to the size of abnormalities. As a key step towards developing a universal abnormality detector, it will be necessary to understand the relationship between different, co-occurring abnormalities, including those with overlapping radiologic appearance such as pneumonia versus atelectasis. Although data from this study contained the typical case mix of a health system, it would also be useful to study the effect of class balance and performance. It will also be advantageous to augment the training data with less commonly encountered abnormalities and with variations in acquisition protocols, such as scans acquired without intravenous contrast material or with lower radiation doses. These additions will enrich the training and improve the generalization of the system. The CT reports and volumes used in this study will be made publicly available pending deidentification and institution approval.

Overall, weak supervision offers a number of general advantages. By using automated label extraction, a vast data set was efficiently annotated from a large health system, which in turn enabled the development of image classifiers for multiple organs and multiple diseases. Such a scalable approach can aid in the process of curating large datasets for developing deep learning models by reducing the need for manual labeling.

## ACKNOWLEDGMENTS

We are grateful for helpful discussions with and data collection by Ricardo Henao PhD, Songyue Han, Khrystyna Faryna MSc, James Tian MD, Mark Kelly MD, Ehsan Abadi PhD, and Brian Harrawood. This work was funded in part by developmental funds of the Duke Cancer Institute as part of the NIH/NCI P30 CA014236 Cancer Center Support Grant, the Center for Virtual Imaging Trials NIH/NIBIB P41-EB028744, and a GPU equipment grant from Nvidia Corp.
.

## CODE AND DATA:

All models' weights, initial hyper-parameters, and code are publicly available at

https://gitlab.oit.duke.edu/railabs/LoGroup/multi-label-weakly-supervised-classification-of-body-ct.

The 13,667 CT volumes and labels will also become publicly available pending institutional approval.

**Table 1.** Number of Patients and Volumes for Each Organ System.

| Label | All* | Training* | Validation* | Testing* |
|---|---|---|---|---|
| A. Lungs and pleura | | | | |
| All patients | 4639 (5044) | 3111 (3514) | 757 (759) | 771 (771) |
| Atelectasis | 1622 (1758) | 1088 (1224) | 283 (283) | 251 (251) |
| Nodule | 1455 (1628) | 918 (1091) | 241 (241) | 296 (296) |
| Emphysema | 1089 (1194) | 710 (815) | 186 (186) | 193 (193) |
| Effusion | 1330 (1465) | 914 (1049) | 211 (211) | 205 (205) |
| No apparent disease | 1307 (1396) | 894 (983) | 204 (204) | 209 (209) |
| B. Liver and gallbladder | | | | |
| All patients | 3835 (4383) | 2536 (3081) | 650 (650) | 649 (652) |
| Hepatobiliary calcification | 888 (1088) | 606 (806) | 138 (138) | 144 (144) |
| Lesion | 1430 (1716) | 958 (1244) | 248 (248) | 224 (224) |
| Dilatation | 537 (628) | 357 (448) | 93 (93) | 87 (87) |
| Fatty | 958 (1108) | 631 (781) | 161 (161) | 166 (166) |
| No apparent disease | 1063 (1123) | 725 (785) | 172 (172) | 166 (166) |
| C. Kidneys and ureters | | | | |
| All patients | 4770 (4770) | 3274 (3274) | 747 (747) | 749 (749) |
| Stone | 1188 (1188) | 823 (823) | 191 (191) | 174 (174) |
| Lesion | 1685 (1685) | 1212 (1212) | 235 (235) | 238 (238) |
| Atrophy | 683 (683) | 473 (473) | 116 (116) | 94 (94) |
| Cyst | 1449 (1449) | 1032 (1032) | 183 (183) | 234 (234) |
| No apparent disease | 1146 (1146) | 783 (783) | 169 (169) | 194 (194) |

Note.— Each volume may contain one or more organs and/or disease labels. The "no apparent disease" classification for patients and volumes indicates that there were no abnormalities. A total of 19,255 labels from 13,667 volumes were from 12,092 distinct patients (mean age 57 ± 18 years; 6172 [51%] women).

* Values shown are number of patients (number of volumes).

**Table 2:** Rule-based Algorithm Performance for 2,158 Radiology Reports Across 15 Labels Using Manually Obtained Ground Truth.

| Label | No. Positive | Accuracy | F-score |
|---|---|---|---|
| Lungs and pleura | | | |
|   Atelectasis | 251 | 98% | 0.97 |
|   Nodule | 296 | 92% | 0.89 |
|   Emphysema | 193 | 99% | 0.98 |
|   Effusion | 205 | 98% | 0.97 |
|   No apparent disease | 209 | 98% | 0.96 |
| Liver and gallbladder | | | |
|   Hepatobiliary calcification | 144 | 96% | 0.91 |
|   Lesion | 224 | 95% | 0.92 |
|   Dilatation | 87 | 98% | 0.92 |
|   Fatty | 166 | 98% | 0.96 |
|   No apparent disease | 166 | 96% | 0.93 |
| Kidneys and ureters | | | |
|   Stone | 174 | 93% | 0.85 |
|   Lesion | 238 | 91% | 0.86 |
|   Atrophy | 94 | 99% | 0.97 |
|   Cyst | 234 | 96% | 0.94 |
|   No apparent disease | 194 | 96% | 0.92 |

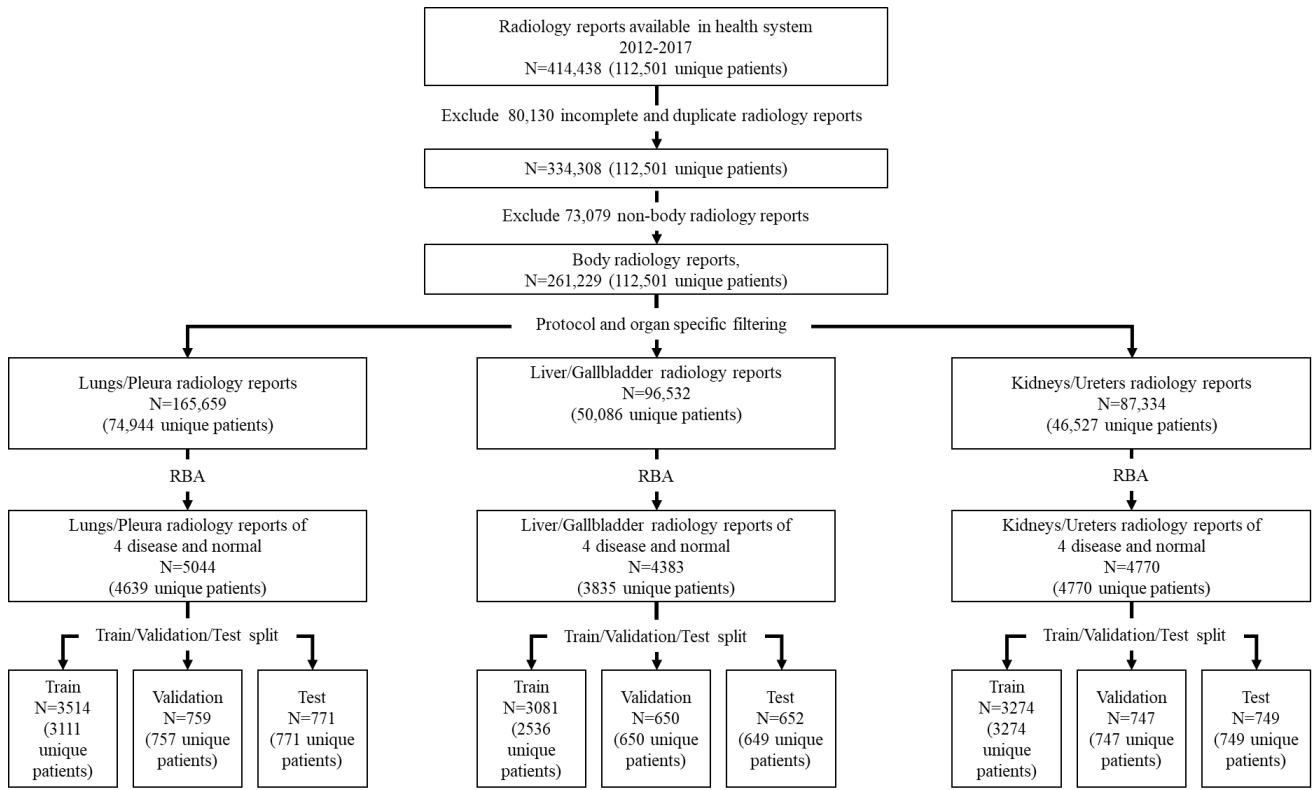

Figure 1. Flowchart shows patient selection. From 414,438 radiology reports performed between 2012 to 2017, a subset of 261,229 reports was selected after excluding incomplete, duplicate, and non-body reports. Those reports were then filtered by protocol and organ-specific topology for the RBA analysis to yield 165,659 reports from 74,944 patients for lungs and pleura; 96,532 reports from 50,086 patients for liver and gallbladder; and 87,334 reports from 46,527 patients for kidneys and ureters. Note that the sum of reports for each organ system does not correspond to the total number of body radiology reports because a single patient could have multiple findings across multiple organ systems. The RBA labeled 5,044, 4,383, and 4,770 reports positive for 4 common diseases or no apparent disease for lungs and pleura, liver and gallbladder, and kidneys and ureters, respectively. Finally, the labeled reports were divided into train, validation, and test by patient. N=number of reports.

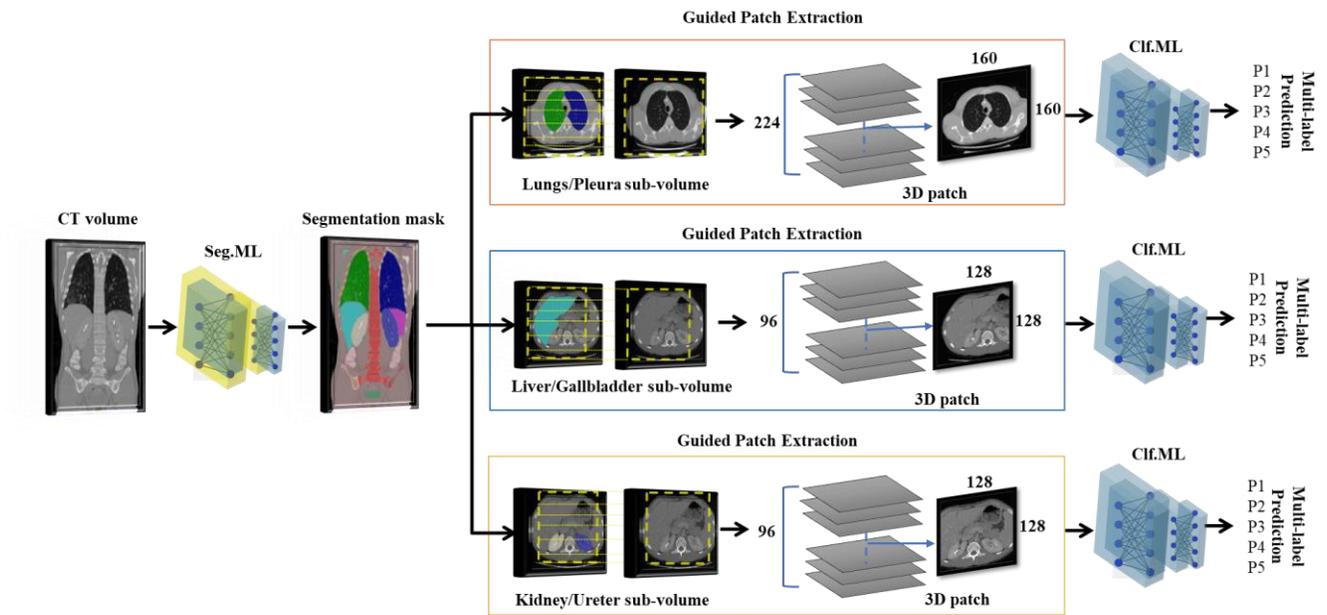

2a.

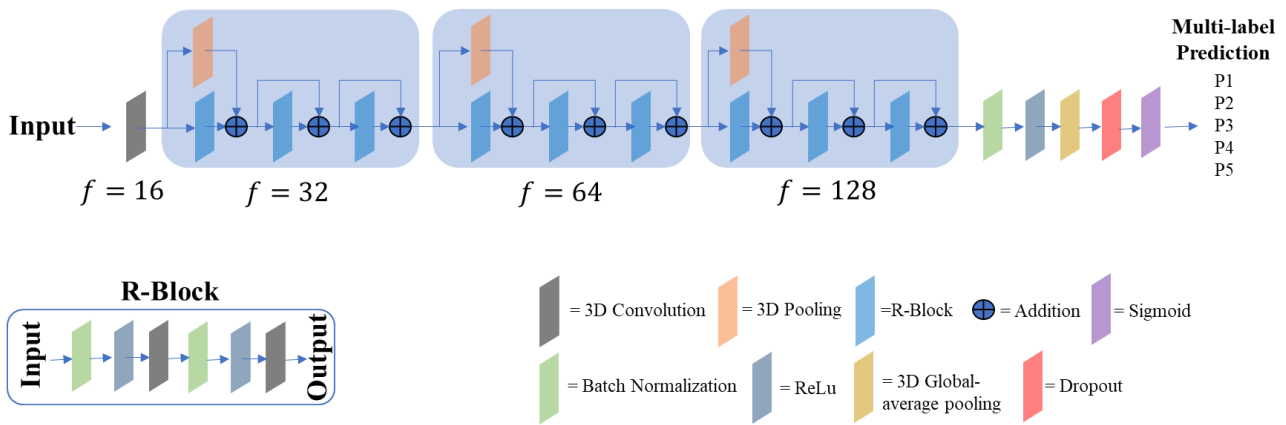

2b.

Figure 2. Overall classification framework and functional architectures. **(A)** The Classification framework includes two networks denoted as Seg. The segmentation module is a DenseVNet (19) network, producing major organ masks such as lungs, liver, and kidneys. Expanded three-dimensional (3D) patches were extracted, which were centered and offset for each organ based on the segmentation masks. Finally, extracted 3D patches were input into the classification module for the final classification score. **(B)** The classification module is a 3D Resnet-like model with 3 R-Blocks in each resolution. Number of filters is denoted as $f$. Final output is a tensor of probabilities for desired multiple labels (diseases). Clf.ML = classification module, ML = segmentation module

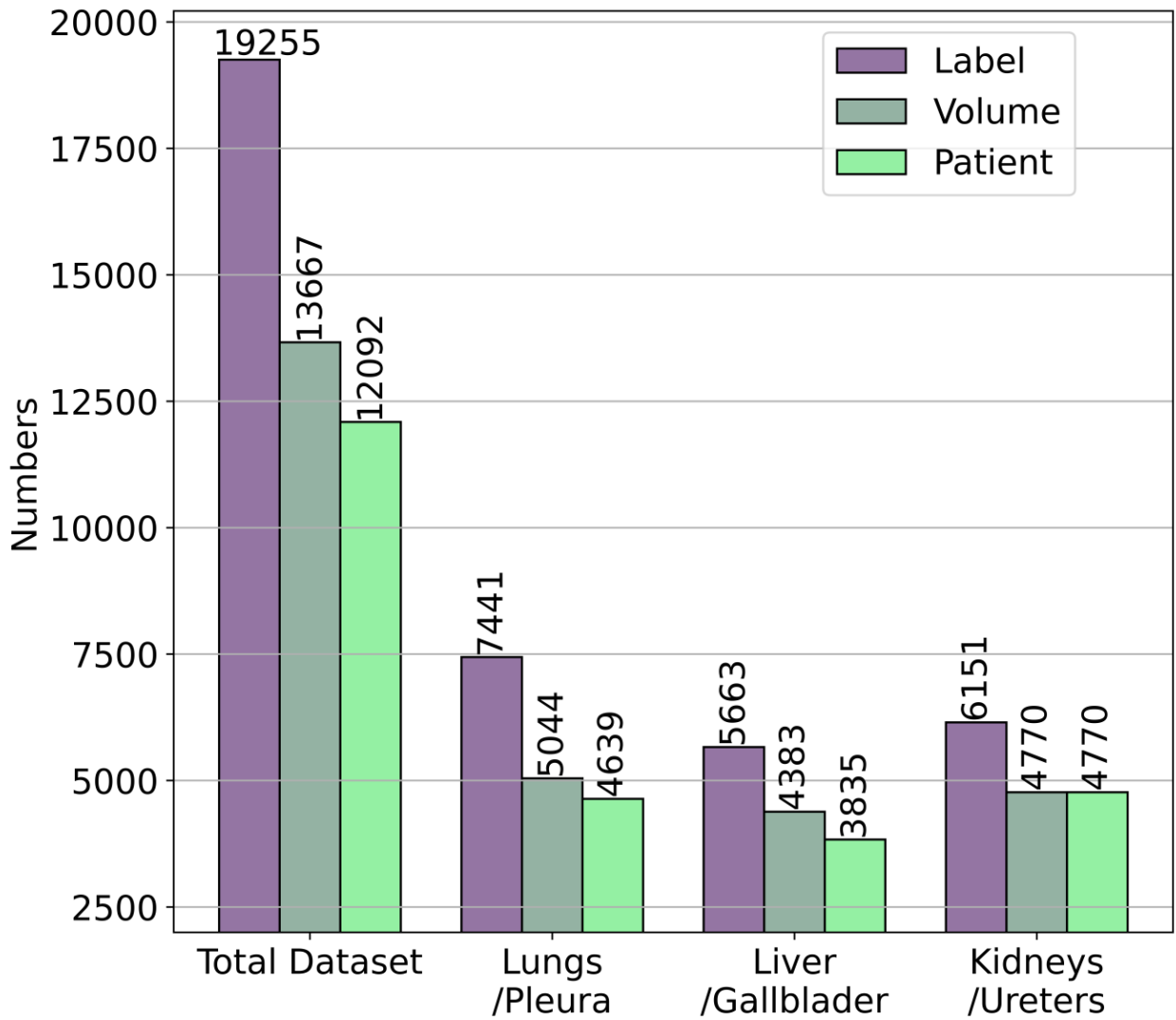

Figure 3. Distribution among three organ systems of 19,255 volume-level labels from 13,667 CT volumes of 12,092 distinct patients.

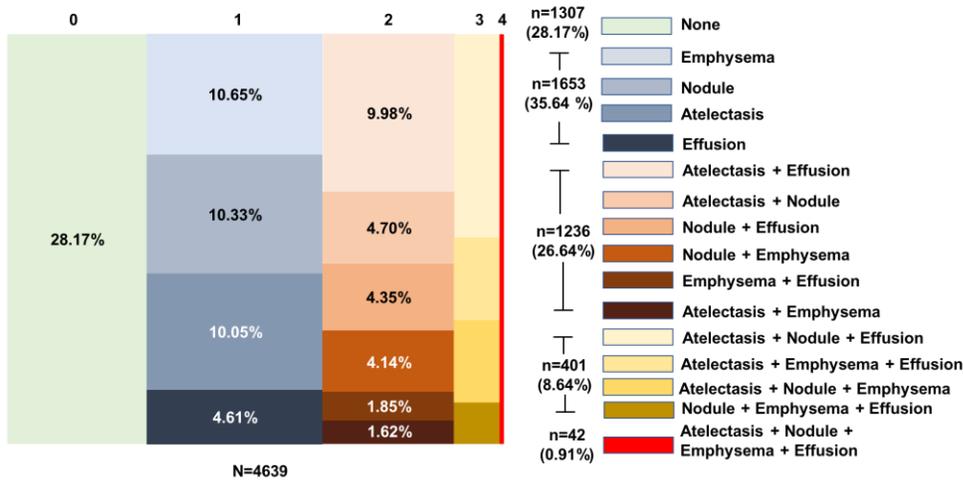
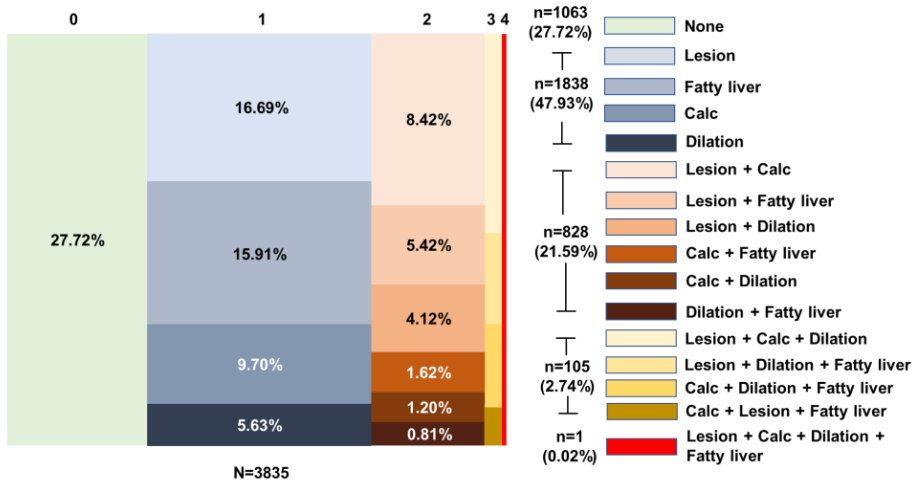
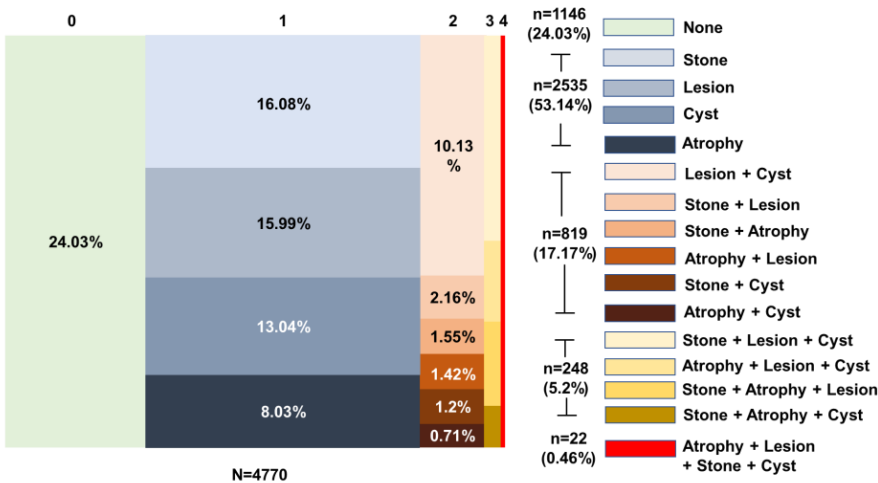

Figure 4. For the (A) lungs and pleura, (B) liver and gallbladder, and (C) kidneys and ureters, tree maps show occurrence and co-occurrence of targeted diseases among distinct patients. Column headers 0, 1, 2, 3 and 4 represent the number of abnormalities within a patient. Area of any column or cell corresponds to frequency for those combination(s) of diseases. For example, in Fig. 4a, column "1" atelectasis (10.05%) and column "2" atelectasis + effusion (9.98%) have approximately the same area. N= number of distinct patients per organ system, n= number of distinct patients belonging to that (co-)occurrence, Calc= hepatobiliary calcification, None= no apparent disease. Percentage of n/N is noted.

5a.

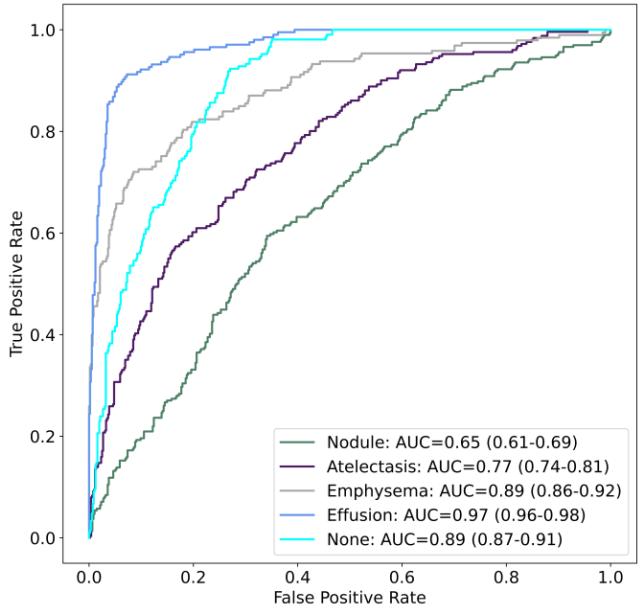

5b.

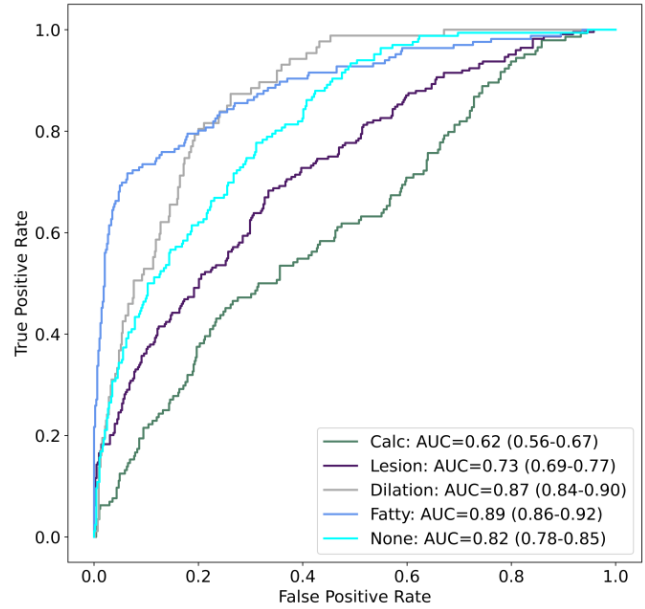

5c.

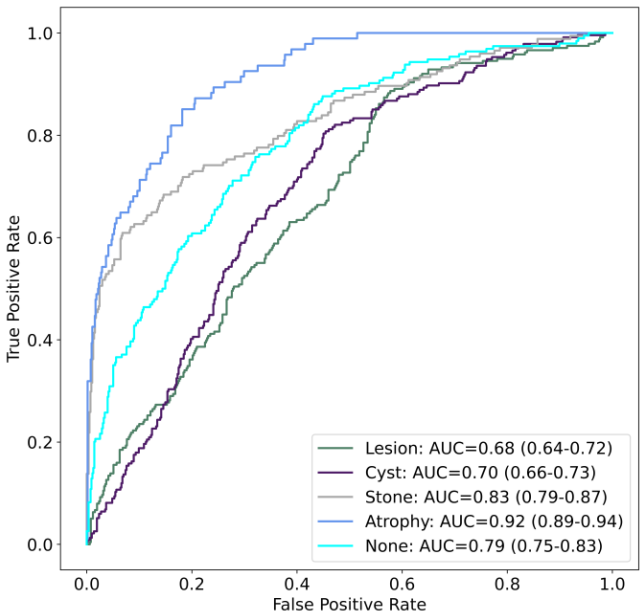

Figure 5. Area under the receiver operating characteristic curve on the test dataset for multi-label classification of CNNs for (A)lungs/pleura, (B) liver/gallbladder, and (C) kidneys/ureters. Numbers in parentheses are represent 95% confidence intervals. Calc= hepatobiliary calcification, None= no apparent disease.

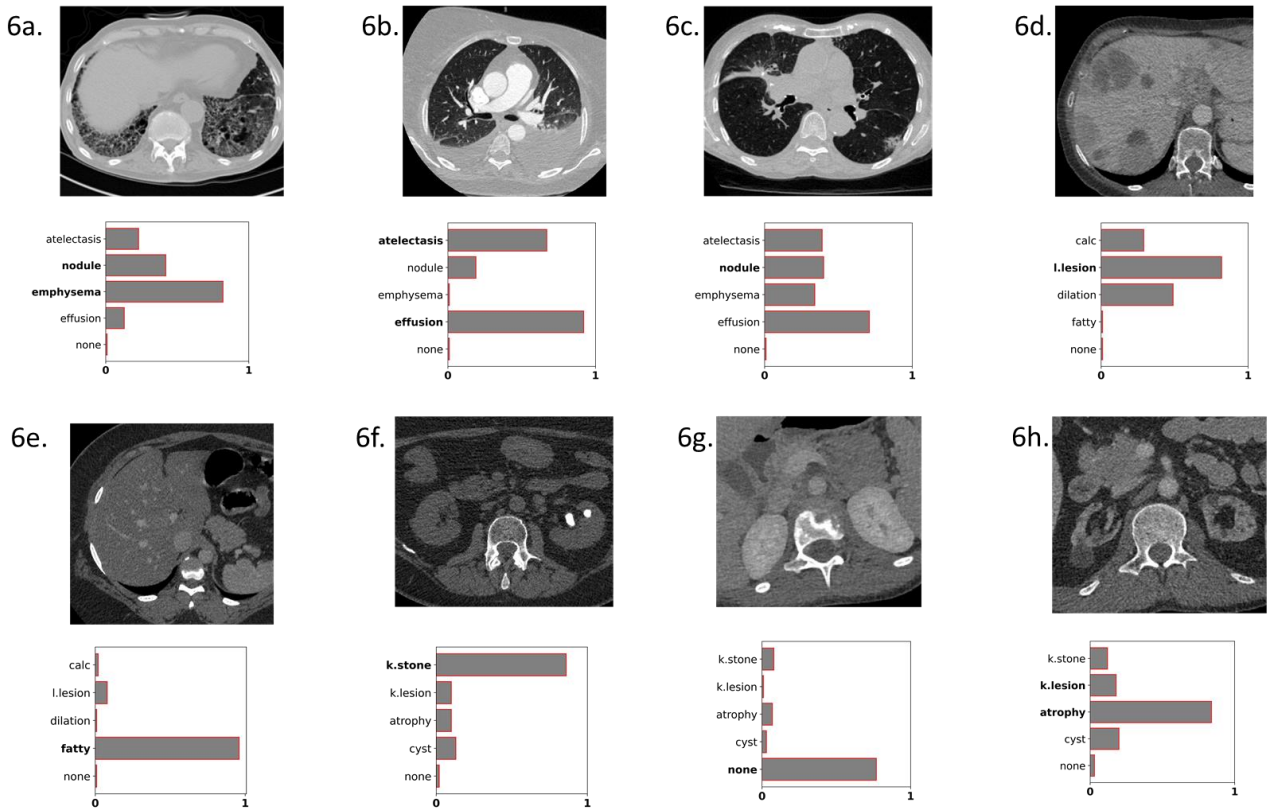

Figure 6. Organ-specific multi-label model predictions relative to ground truth. Each example shows a transverse reconstruction and associated multi-label model outputs for the three-dimensional sub-volume patch. Ground truth labels are listed in bold text, and predicted probabilities indicated by the gray bars. calc= hepatobiliary calcification, l.lesion= liver/gallbladder lesion, k.stone=kidneys/ureters stone, k.lesion=kidneys/ureters lesion, none= no apparent disease. (A) Shows a correct prediction of nodule and emphysema with high probabilities for those two ground truth labels. and low probabilities for the other three labels. (**B**) Shows a correct prediction of atelectasis and effusion with high probabilities for those two ground truth labels and low probabilities for the other three labels. (C) Shows an incorrect prediction of effusion with high probability compared to the ground truth label probability of nodule. (D) Shows a correct prediction of liver lesion with high probability ground truth label and a high incorrect probability for liver dilation. (E) Shows a correct prediction of fatty with high probability for the ground truth label and low probabilities for the other four labels. (F) Shows a correct prediction of kidneys stone with high probability for the ground truth label and low probabilities for the other four labels. (G) Shows a correct prediction of kidneys with no apparent disease and with high probability for the ground truth label and low probabilities for the other four labels.

(H) Shows a correct prediction of kidneys atrophy with high probability for the ground truth label and an incorrect false negative for kidneys stone with low probability for the ground truth label.